\documentclass[letterpaper, 10 pt, conference]{ieeeconf}  

\IEEEoverridecommandlockouts                              

\usepackage{times}
\usepackage{epsfig}
\usepackage{graphicx}
\usepackage{amsmath}
\usepackage{amssymb}
\usepackage{tikz}
\usepackage{bigfoot}
\newcommand{\tikzcircle}[2][red,fill=red]{\tikz[baseline=-0.5ex]\draw[#1,radius=#2] (0,0) circle ;}%

\newcommand\blfootnote[1]{%
	\begingroup
	\renewcommand\thefootnote{}\footnote{#1}%
	\addtocounter{footnote}{-1}%
	\endgroup
}

\makeatletter
\let\NAT@parse\undefined
\makeatother
\usepackage[pagebackref=true,breaklinks=true,colorlinks,bookmarks=false]{hyperref}

\title{\LARGE \bf
What Can be Seen is What You Get:\\Structure Aware Point Cloud Augmentation
}

\author{
	Frederik Hasecke$^{1,2}$,  Martin Alsfasser$^2$ and Anton Kummert$^1$
	
\thanks{$^{1}$ Bergische Universit\"at Wuppertal
        {\tt\small \{frederik.hasecke, kummert\}@uni-wuppertal.de}}
\thanks{$^{2}$ Aptiv Services Deutschland
        {\tt\small \{frederik.hasecke, martin.alsfasser\}@aptiv.com}}%
}

\begin{document}

\maketitle
\thispagestyle{empty}
\pagestyle{empty}

\begin{abstract}
	To train a well performing neural network for semantic segmentation, it is crucial to have a large dataset with available ground truth for the network to generalize on unseen data.
	In this paper we present novel point cloud augmentation methods to artificially diversify a dataset.
	Our sensor-centric methods keep the data structure consistent with the lidar sensor capabilities. Due to these new methods, we are able to enrich low-value data with high-value instances, as well as create entirely new scenes.
	We validate our methods on multiple neural networks with the public SemanticKITTI \cite{behley2019semantickitti} dataset and demonstrate that all networks improve compared to their respective baseline. 
	In addition, we show that our methods enable the use of very small datasets, saving annotation time, training time and the associated costs.

\end{abstract}


\blfootnote{}
\blfootnote{\copyright 2022 IEEE. Personal use of this material is permitted. Permission from IEEE must be obtained for all other uses, in any current or future media, including reprinting/republishing this material for advertising or promotional purposes, creating new collective works, for resale or redistribution to servers or lists, or reuse of any copyrighted component of this work in other works.}
\section{INTRODUCTION}
\label{section:intro}
Semantic segmentation for lidar data is the task to assign a certain class label to every point in a point cloud. 
In this work we address inherent bias contained in the training data and try to reduce the impact through augmentation methods.
The main issue with recording training data in the real world is the imbalance of classes. While most roads are populated with cars it can be rare to come across a motorcyclist in some areas for instance. This imbalance also shows in public datasets \cite{behley2019semantickitti}\cite{caesar2020nuscenes}. The SemanticKITTI \cite{behley2019semantickitti} dataset provides an overview of the total number of points per class. The amount of car points is three orders of magnitude higher than that of motorcyclists. 

Despite a number of successful recent approaches for augmentation methods to overcome imbalance and overfitting issues in the two-dimensional image domain \cite{devries2017improved}\cite{zhang2017mixup}\cite{yun2019cutmix}, the extension of these concepts into the three-dimensional lidar domain creates a clear disconnect between original data and augmented data \cite{chen2020pointmixup}\cite{zhang2021pointcutmix}\cite{nekrasov2021mix3d}. The structure of the data and the field of view of the sensor are changed to a great extent by these methods, which makes the use of e.g., structure-dependent networks\cite{milioto2019rangenet++}\cite{cortinhal2020salsanext} impossible.

\begin{figure}
	\centering
	\includegraphics[width=0.45\textwidth]{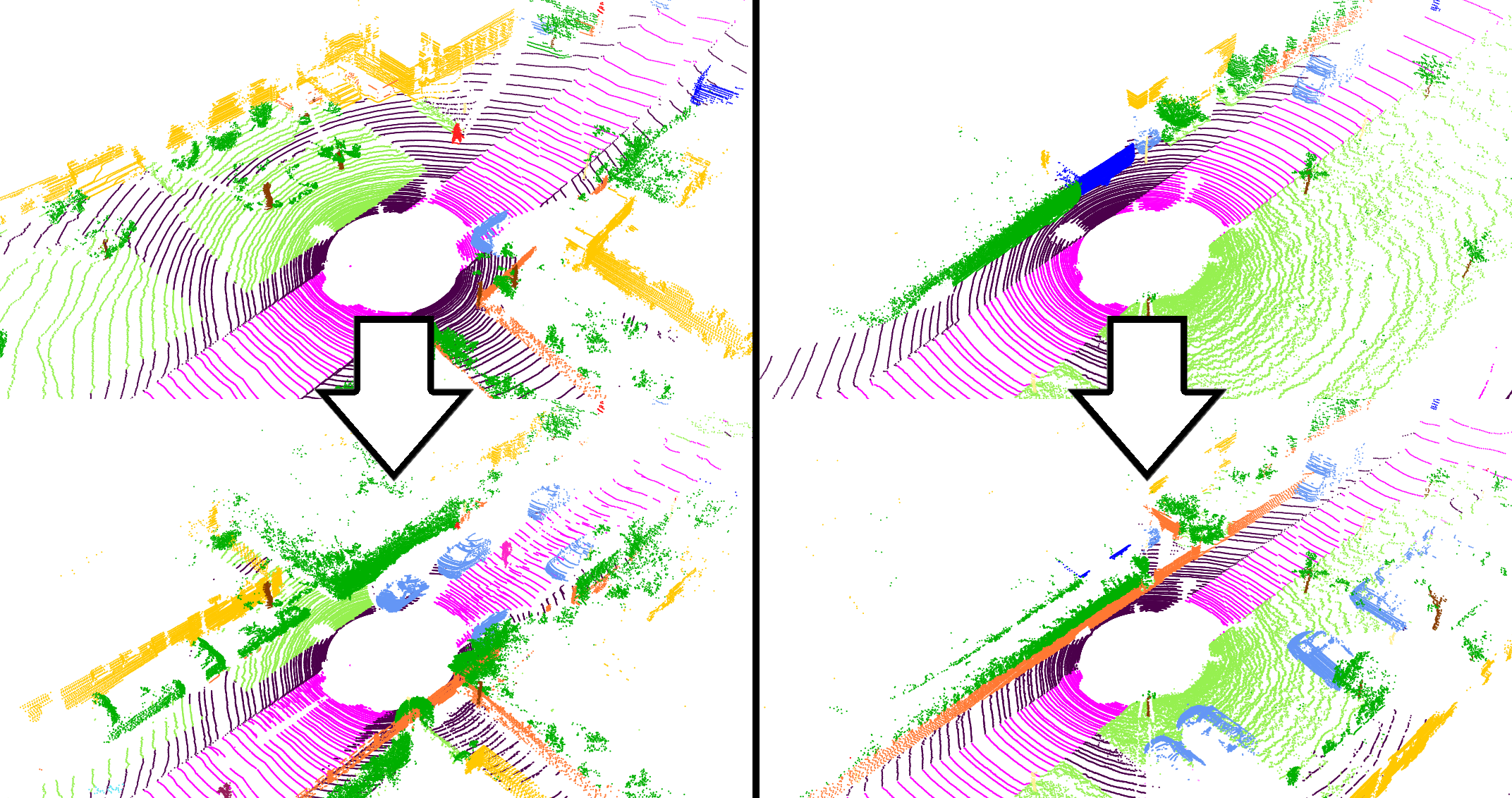}
	\caption{\textbf{Two lidar point clouds enriched via the structure aware point cloud augmentation methods shown in this paper.} The original lidar point clouds (\textit{top}) are augmented with parts of other point clouds to create entirely new scenes (\textit{bottom}) for neural networks. Best viewed in color}
	\label{fig:result}
\end{figure}

We aim to balance out the class bias of the data without overfitting to the underrepresented classes. At the same time our goal is to maintain the structure of the original raw lidar data, while augmenting the content. We do so by introducing two new augmentation modules for lidar data. Two generated example point clouds of our augmentation pipeline can be seen in \textsc{Figure} \ref{fig:result}. Our augmentation pipeline consists of two new modules we can add to any neural network for semantic segmentation of lidar data; they are not limited to a certain kind of network. Our methods use the lidar sensors operation principle as the core assumption to augment the data in a structurally conservative way. Our first method injects additional objects into a lidar point cloud. We remove the occluded points in a point-wise comparison of the injection to the target point cloud to keep a realistic and structurally correct point cloud. The second method we propose in this work is a complete fusion between two point clouds to generate a third out of various parts of the two parent point clouds. The result is a newly generated point cloud, which shows the same structural features as an original recorded lidar point cloud. We evaluate our augmentation pipeline on current state-of-the-art networks for semantic segmentation of outdoor automotive lidar data. To demonstrate the model independence of our methods we use networks from the three most common approaches. We use KPConv\cite{thomas2019kpconv} for the point-wise models, Cylinder3D\cite{zhou2020cylinder3d} for the voxel-based models and for the range-image-based models RangeNet++\cite{milioto2019rangenet++} and SalsaNext\cite{cortinhal2020salsanext}. Our main contributions of this work are therefore as follows:
\begin{itemize}
	\item We introduce a novel point cloud injection method which preserves the original lidar structure.
	\item We propose a new point cloud fusion method that merges two lidar point clouds while preserving the physical structure to generate new scenes, which are indistinguishable from original raw lidar data
	\item We investigate the impact of our augmentation modules on multiple open source networks with a comprehensive ablation study 
	\item We achieve competitive results on the SemanticKITTI dataset and demonstrate a significant boost in performance over the original networks
\end{itemize}

\section{Related Works}
\label{section:related}
Neural networks are considered well performing when they generalize to unseen data. When deploying a trained network, one expects a comparable result to that of the validation data that is used to tune the network.
Due to the aforementioned overfitting to training data this expectation might not be met. There are various approaches to address this problem. In the following we divide these approaches into model-related and data-related measures. 

The first category of overfitting counter measures consists of methods such as regularization \cite{hanson1988comparing}\cite{weigend1991generalization} and dropout \cite{srivastava2014dropout}. The former technique discourages the model from overfitting by imposing a penalty on complexity. The later randomly "drops" parts of the network to prevent co-dependencies among parts of the network. 

The second category represents measures that are applied directly to the training data. The overarching goal is to continuously change the data so that overfitting can be prevented. This can also include methods such as over-sampling rare classes \cite{ling1998data}, expanding the training data with other datasets \cite{rame2018omnia} and pretraining the network on other data domains \cite{brekke2019multimodal}.

In the point cloud area however data augmentation primarily relates to the direct modification of the existing training data. We distinguish between global augmentations, which are applied to an entire training sample, local augmentations which are applied to individual objects and, ultimately, context augmentations, where samples of the training data are mixed with other data samples.

\subsection{Global Augmentations}
\label{section:related:global_augmentations}
Lidar point clouds as well as point cloud from other sensor types are usually augmented by applying random translations, rotations, flips and point drops. These techniques show improved generalization of neural networks due to a seemingly larger training dataset \cite{liong2020amvnet}\cite{yan2020sparse}. 

\subsection{Local Augmentations}
\label{section:related:local_augmentations}
In two-dimensional image domain, different variations of copy and paste augmentations \cite{yun2019cutmix} have been successfully used in various applications. These methods cut out parts of an image and paste them onto another image. This approach is also found in the three-dimensional lidar area. Previous works have already shown the benefits of injecting underrepresented classes in object detection tasks into training points. The most notable example is \cite{yan2018second} in which Yan et al. create a database that contains the cut-out point clouds of rare objects using the three-dimensional cuboid annotation around the objects. During training these objects are sampled from the database and injected into the point cloud at random locations. To ensure physically possible locations the injected objects are compared in bird's-eye-view with other objects to avoid collisions \cite{yan2018second}. The work of \cite{alsfasser2020exploiting} additionally removed all points behind the cuboid of the injected object in polar coordinates to remove any possible overlap with existing objects, thus sidestepping the collision issue mentioned by \cite{yan2018second}. In a more recent publication 
\cite{hu2021pattern} applied a sub-sampling of the original sampled injection object along the scanlines and between scanlines to extend the distance of injections in the target point cloud, while keeping a similar structure of the lidar scanlines. The authors of \cite{zhu2021vpfnet} use a multi-modal approach, in which they use an instance segmentation network to sample \textit{"foreground"} instances in the camera image. These are projected into the lidar points to remove occluded points. Lately, injection methods have also been mentioned in lidar segmentation publications \cite{xu2021rpvnet}\cite{zhou2021panoptic} which are described similarly to the method in \cite{yan2018second}. The most comparable approach to ours is \cite{fan2021rangedet} in which the authors apply a \textit{Copy-Paste} algorithm \cite{yun2019cutmix} to the range image to inject cars in front of other objects, but cancel injections which are behind other objects (e.g., walls). 

Our method is the first to apply a point-wise occlusion competition between target and injection points, thus removing occluded points from the target point cloud but also the injection objects according to their distance from the lidar sensor. This ensures that the lidar sensor structure is retained and thus the augmented data cannot be distinguished from real raw data.  


\subsection{Context Augmentations}
\label{section:related:fusion}
Augmenting the context of an entire scene is another approach to prevent overfitting of a network. Object recognition models and models for semantic segmentation tend to perceive individual objects together with their environment. In doing so, the network can place an excessive focus on the environment, which can change on unseen data in a different context \cite{rosenfeld2018elephant}\cite{nekrasov2021mix3d}. Context augmentation is used to minimize such connections between objects and their context by using individual objects in other scenes (see \textsc{Section} \ref{section:related:local_augmentations}) or by mixing entire scenes.

In the image domain we find several examples in which parts of pictures are completely removed \cite{devries2017improved}, parts are cut out and pasted onto other pictures \cite{yun2019cutmix}, or several pictures are blended together in order to minimize the contextual dependence of the image content \cite{zhang2017mixup}.

This type of augmentation is also used in the three dimensional domain. The approach presented in \cite{chen2020pointmixup} blends two point clouds of objects to generate new examples. The work in \cite{zhang2021pointcutmix} extends the approach of  \cite{yun2019cutmix} into the three-dimensional domain and cuts and pastes parts of one object point cloud into another to create new mixes and enhance the model robustness against point attacks. The authors of \cite{nekrasov2021mix3d} overlay two full scenes, to create a new combination scene. The authors show that out-of-context mixing reduces overfitting to the training set, which leads to a better generalization on unseen data.

Our method for context augmentation is the first to use a sensor-centric approach to the fusion of two point cloud scenes. Thus, opposed to all previous mentioned methods, we keep the lidar sensor structure intact and provide fused scenes which are structurally indistinguishable from real raw data.

\section{Method}
\label{section:method}
In this paper we propose lidar structure aware point cloud augmentation methods. Lidar sensors, such as the one used in the SemanticKITTI \cite{behley2019semantickitti} dataset, consist of vertically stacked light emitting and receiving sensor modules. These rotate around a shared central axis. All range detections along a single revolution are streamed into a single point cloud file, which represents the surrounding of the sensor. Due to the distinct vertically stacked laser sensor modules the point cloud exhibits the characteristic scanline shape, in which ring-like patterns can be seen as shown in \textsc{Figure} \ref{fig:result}. 

The amount of lidar rings is directly related to the amount of vertically stacked lidar sensor modules of the scanner. The number of individual range measurements of each ring on the other hand depends on the recording frequency of the sensor. 

\subsection{Data Augmentation}
\label{section:method:augmentation}
Most global augmentation methods, such as the ones mentioned in \textsc{Section} \ref{section:related:global_augmentations} keep the previously mentioned lidar structure intact.
By globally rotating, flipping or slightly moving the point cloud around the sensor origin, these methods ensure, that the scanlines and the line-wise point distances are kept the same relative to each other. We use these well known concepts in our augmentation pipeline for a global augmentation of the given point cloud but extend the usage to single objects and mixed scenes.

\subsection{Structure Aware Point Cloud Injection}
\label{section:method:injection}
Our injection method builds up on the concepts presented in \textsc{Section} \ref{section:related:local_augmentations}, but extends the sampling and the injection into the range image domain based on the recording method of rotating lidar sensors.
The range image is a very close representation of what the lidar sensor is able to capture. We use the semantic point-wise labels to extract rare objects from the training data without including the ground, unlike the methods mentioned in \textsc{Section} \ref{section:related:local_augmentations}. We store all sampled objects in a database for optimized access.
\begin{figure}
	\centering
	\includegraphics[width=0.4\textwidth]{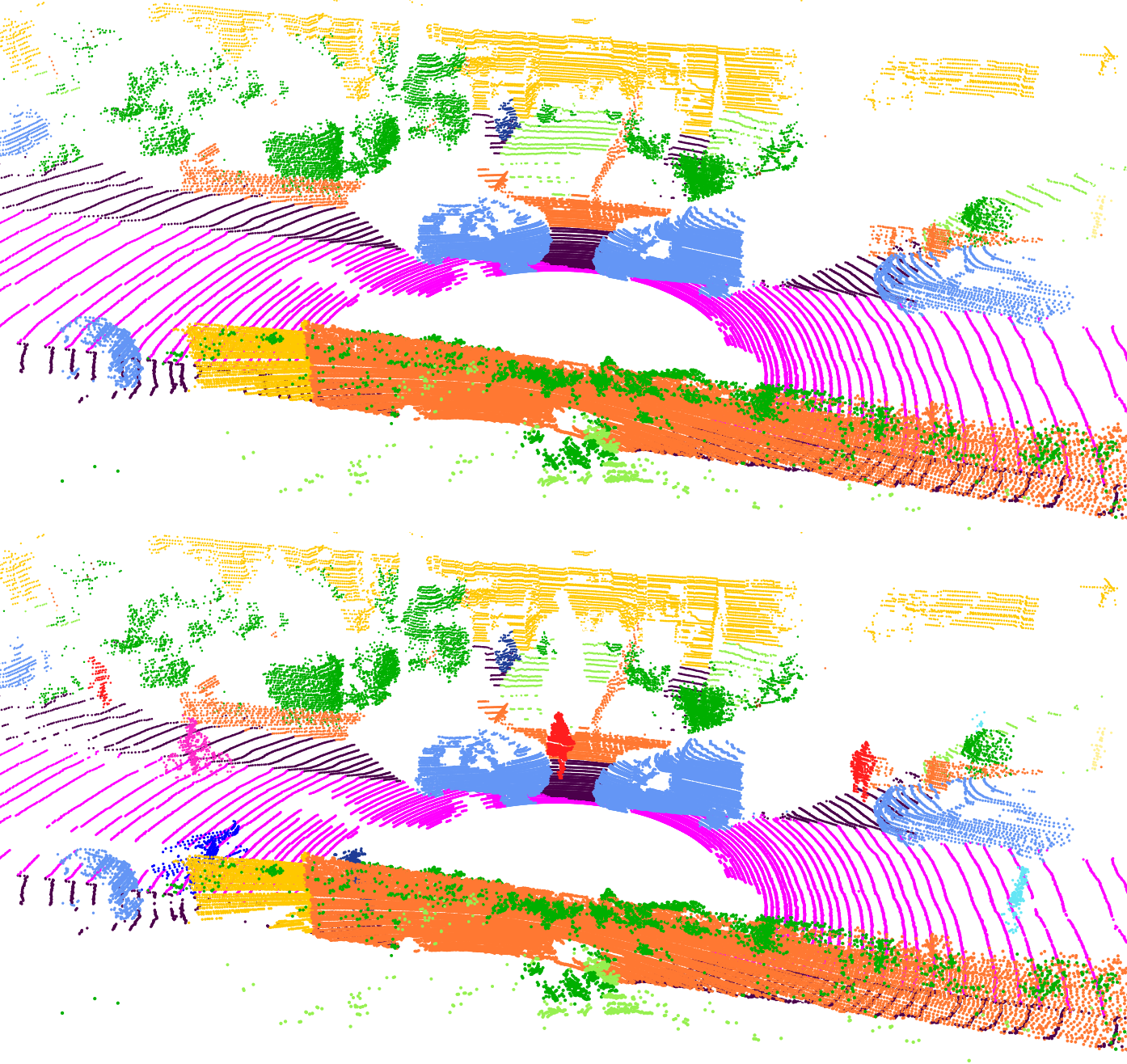}
	\caption{\textbf{Additional instances injected into a scene.} \textit{Top}: Original lidar scene. \textit{Bottom}: Our augmented scene enriched with additional 
		\tikzcircle[black, fill=red]{2.5pt}"$Pedestrians$", 
		\tikzcircle[black, fill=cyan]{2.5pt}"$bicycles$",
		\tikzcircle[black, fill=magenta]{2.5pt}"$bicyclists$" and
		\tikzcircle[black, fill=blue]{2.5pt}"$other-vehicles$". Best viewed in color}
	\label{fig:inject}
\end{figure}
In the injection step we sample injection objects from our database and apply global augmentations (flips, rotations and point drops) to the object. As these augmentations retain the original lidar structure, we can directly project the injection object into the range image domain. We compare each pixel in the range image of the to-be-injected object with the corresponding pixels in the target point cloud: The point closer to the sensor wins the competition and remains in the new point cloud, the farther point is removed. The advantage of this method is, that we inject objects e.g., behind street lights, and remove the occluded points in the injection object instead of preventing the injection which is different to the methods mentioned in \textsc{Section} \ref{section:related:local_augmentations}. With the same technique we remove the points behind the injected object to imitate the lidar shadow behind the object.

These two steps - sampling and injecting - enable us to enrich any training point cloud with additional objects, especially of rare classes, boosting the performance of neural networks on these. We also keep the structure of the point cloud consistent with what the lidar sensor can capture. Therefore, we create point clouds closer to the validation and test data than other naive injection methods, which would create a noticeable difference between training and test data, as well as overlapping point clouds in the sphere coordinate view.
\textsc{Figure} \ref{fig:inject} shows the direct comparison of a training point cloud and the same point cloud augmented by our structure aware point cloud injections.

In order to inject underrepresented classes, out approach compares the present class distribution in the target point cloud with a desired distribution. Additionally we define a parameter for a maximum number of injections and start the process by randomly choosing an injection class. If the given class is already present in the target point cloud, in a proportion larger than the desired amount, we switch to a different randomly chosen injection class. We continue to randomly sample, check and inject objects into the target point cloud until we either reach the desired class distribution, or the maximum number of injections.

\subsection{Structure Aware Point Cloud Fusion}
\label{section:method:fusion}
\begin{figure*}
	\centering
	\includegraphics[width=0.9\textwidth]{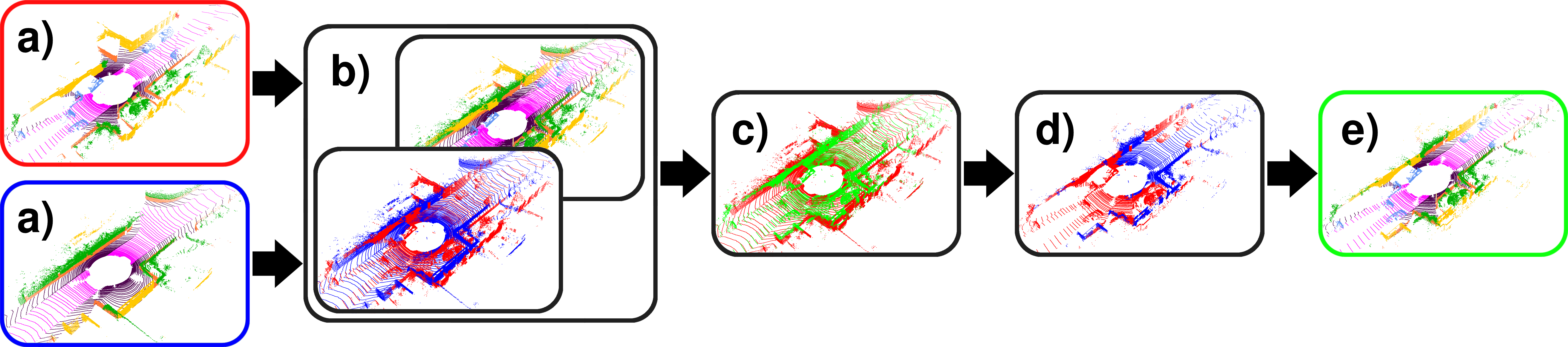}
	\caption{\textbf{Visualization of our point cloud fusion method.} \textbf{a)} Two separate independent point cloud from the training set. \textbf{b)} Both point clouds overlayed on top of each other. \textbf{c)} Range competition of both point clouds: The green points are closer to the lidar sensor. \textbf{d)} Fused point cloud of the closer points of both parent point clouds. \textbf{e)} Final fused point cloud, exhibiting parts of each parent point cloud. Best viewed in color.}
	\label{fig:fusion}
\end{figure*}
The second augmentation module in this work goes beyond the injection of single objects into an otherwise unchanged point cloud. We load a second training point cloud and apply global structure-maintaining augmentations to it. To ensure a valid point cloud, we rotate the second cloud in increments of the horizontal sensor resolution and prevent a translation of the points. We also do not scale the point cloud, but allow random flips in the x and y direction. These restrictions keep the structure of the second point within the lidar sensors recording capabilities.

Based on the range competition technique mentioned in \textsc{Section} \ref{section:method:injection}, we project both point clouds into the range image domain and compare both pixel-wise. We keep the closer range measurement and discard the farther one to generate a new scene, which results in a structurally as well as semantically valid point cloud.
We use a rotation limitation, to keep the rotation of the sensor in a $\pm 10^\circ$ angle, to keep the probable direction along the street valid. Most recordings of automotive lidar data have a road at the front and back, while the sides are obstructed by e.g., houses and parking cars, we want to keep this orientation for the generated fused point cloud. \textsc{Figure} \ref{fig:result}  and \textsc{Figure} \ref{fig:fusion} e) show examples of these generated scenes. The global augmentations (i.e., rotations, flips and point drops) applied to the second point cloud lower the likelihood of ever merging the same point clouds in the same arrangement, to a negligibly small probability. \textsc{Figure} \ref{fig:fusion} visualizes 
the process to generate a point cloud consisting of parts of the two parent point clouds. While the lidar data structure is preserved, there is a risk of objects from both point clouds merging and thus causing an unrealistic representation. It can also happen that nonsensical scenes are created, such as a freeway guardrail that blocks a house entrance, or a tree that protrudes from the roof of a car. However, we've noticed that these nonsensical, out-of-context blendings improve networks rather than confuse them.
In order to keep the parameters of the injection module valid, we apply the fusion step before the injection step in our pipeline.

\section{Experiments}
\label{section:experiments}

We added our proposed augmentation methods to the training data-loaders of different neural networks. 
As we augment the raw sensor data we are not limited to a special kind of network, but are instead able to add them to point-based, voxel-based and also range-image-based networks without any issue, as the generated point clouds exhibit the same structure and orientation as real raw data. In \cite{triess2020scan} a method was proposed to remove the influence of the in-frame ego motion compensation applied to the SemanticKITTI \cite{behley2019semantickitti} dataset. We used this method and extended the principle to Sequence 3 of SemanticKITTI by applying the reverse relative transformation matrix of the odometry data to the lidar data.
For all networks we use the same augmentation parameters which are as follows:
\begin{itemize}
	\item 50\% probability to apply global augmentations
	\item 30\% probability to mix with a second random scene
	\item 50\% probability to inject instances
	\item maximum of three injected instances per frame
	\item desired share of 2\% for all injection classes

\end{itemize}
\vspace{5pt}

\textbf{Point-based Networks}\\
We used KPConv \cite{thomas2019kpconv} as an open source point-wise network for our augmentation methods, as it currently is the best performing published point-wise network on the SemanticKITTI \cite{behley2019semantickitti} leaderboard. We used the Pytorch implementation by the original authors\footnote{https://github.com/HuguesTHOMAS/KPConv-PyTorch} with slight modifications to the training parameters to fit the training and validation loop into our GPU, namely we increased the input radius of each ball query to 51 meters and the size of the first sub-sampling grid to 0.2 meters. We left the other parameters of the network as the authors provided them. These hardware based limitations lead to a decline in performance of the network on small classes such as bicycles and pedestrians as can be seen in \textsc{Table} \ref{tab:results} later in this section.\\

\textbf{Voxel-based Networks}\\
As a representative for the currently leading group of networks for the semantic segmentation of lidar point clouds - voxel-based networks - we use the public repository\footnote{https://github.com/xinge008/Cylinder3D} of the Cylinder3D \cite{zhou2020cylinder3d} network, which was published by the original authors. 
All 
training parameters have been left as the authors provided them.\\

\textbf{Range-image-based Networks}\\
For the last group we used two open source networks: Rangenet++ \cite{milioto2019rangenet++}\footnote{https://github.com/PRBonn/lidar-bonnetal} and SalsaNext \cite{cortinhal2020salsanext}\footnote{https://github.com/Halmstad-University/SalsaNext}. These network types are based on the lidar structure of the point clouds and require a consistent structure to function properly. Therefore we used two networks to showcase the structure retention of our augmentation pipeline. We did not change any parameters of these two networks, and used the training scripts as provided by the authors.\\
\begin{table*}[t]
	\caption{\textbf{Results of various Networks with and without our Augmentation Pipeline on the SemanticKITTI Validation Set}}
	\resizebox{\textwidth}{!}{
		\begin{tabular}{c|c|c|c|c|c|c|c|c|c|c|c|c|c|c|c|c|c|c|c|c}
			\hline
			\textbf{Methods} & \textbf{mIoU} & \rotatebox[origin=c]{90}{car} & \rotatebox[origin=c]{90}{bicycle} & \rotatebox[origin=c]{90}{motorcycle}& \rotatebox[origin=c]{90}{truck}& \rotatebox[origin=c]{90}{other-vehicle} & \rotatebox[origin=c]{90}{person} & \rotatebox[origin=c]{90}{bicyclist} & \rotatebox[origin=c]{90}{motorcyclist} & \rotatebox[origin=c]{90}{road} & \rotatebox[origin=c]{90}{parking} & \rotatebox[origin=c]{90}{sidewalk} & \rotatebox[origin=c]{90}{other-ground} & \rotatebox[origin=c]{90}{building}& \rotatebox[origin=c]{90}{fence}& \rotatebox[origin=c]{90}{vegetation}& \rotatebox[origin=c]{90}{trunk}& \rotatebox[origin=c]{90}{terrain}& \rotatebox[origin=c]{90}{pole}& \rotatebox[origin=c]{90}{traffic-sign}\\
			\hline

			RangeNet++\cite{milioto2019rangenet++}        & 52.8          & 91.0 & 25.0 & 47.1 & 40.7 & 25.5 & 45.2 & 62.9 & 0.0 & 93.8 & 46.5 & \textbf{81.9} & 0.2 & 85.8 & 54.2 & \textbf{84.2} & 52.9 & \textbf{72.7} & \textbf{53.2} & \textbf{40.0} \\
			RangeNet++\cite{milioto2019rangenet++} + Ours & \textbf{56.3} & \textbf{91.4} & \textbf{34.1} & \textbf{56.8} & \textbf{57.5} & \textbf{39.7} & \textbf{53.4} & \textbf{66.8} &\textbf{6.2} & \textbf{94.4} & \textbf{49.9} & 81.6 & \textbf{0.4} & \textbf{87.0} & \textbf{58.8} & 83.1 & \textbf{53.0} & 71.6 & 45.3 & 37.8 \\
			\hline
			SalsaNext\cite{alonso20203d}        & 56.9 & 86.7 & 40.7 & 42.0 & \textbf{79.3} & 42.5 & 64.6 & 69.4 & \textbf{0.0} & 94.5 & \textbf{42.6} & \textbf{80.2} & 3.5 & 80.6 & 48.3 & 80.8 & 61.6 & 65.1 & 53.1 & 44.9\\
			SalsaNext\cite{alonso20203d} + Ours & \textbf{61.2} & \textbf{90.4} & \textbf{46.2} & \textbf{58.7} & 66.4 & \textbf{54.1} & \textbf{71.9} & \textbf{80.1} & \textbf{0.0} & \textbf{94.6} & 42.1 & \textbf{80.2} &\textbf{ 3.7} & \textbf{87.4} & \textbf{50.7} & \textbf{86.0} & \textbf{67.1} & \textbf{72.5} & \textbf{62.1} & \textbf{48.4} \\
			\hline
			
			KPConv\cite{thomas2019kpconv}        & 43.3 & 93.7 & \textbf{0.0} & 0.9 & 79.6 & 20.6 & 0.0 & \textbf{0.0}  & \textbf{0.0} & 90.4 & 20.7 & 71.7 & \textbf{0.0} & 89.5 & 54.9 & \textbf{86.2} & 51.7 & \textbf{70.2} & 55.9 & \textbf{37.3} \\
			KPConv\cite{thomas2019kpconv} + Ours & \textbf{45.9} & \textbf{93.9} & \textbf{0.0} & \textbf{12.4} & \textbf{80.9} & \textbf{22.3} & \textbf{32.1} & \textbf{0.0} & \textbf{0.0} & \textbf{90.8} & \textbf{23.5} & \textbf{72.8} & \textbf{0.0} & \textbf{89.6} & \textbf{55.7} & 85.4 & \textbf{54.7} & 68.3 & \textbf{56.6} & 32.3 \\
			
			
			\hline

			Cylinder3D\cite{zhou2020cylinder3d} $\dagger$ & 66.9 & 97.1 & 54.5 & 80.9 & \textbf{85.1} & 70.3 & 76.5 & 92.2 & \textbf{0.0} & 94.6 & 44.8 & 81.2 & \textbf{1.0} & \textbf{90.5} & \textbf{58.7} & 86.6 & \textbf{70.8} & 70.5 & 64.2 & \textbf{51.8} \\
			Cylinder3D\cite{zhou2020cylinder3d} + Ours    & \textbf{67.9} & \textbf{97.3} & \textbf{58.6} & \textbf{85.8} & 84.8 & \textbf{71.4} & \textbf{81.8} & \textbf{92.8} & 0.8 & \textbf{94.9} & \textbf{45.3} & \textbf{81.8} & 0.4 & 90.2 & 58.3 & \textbf{87.6} & 69.0 & \textbf{72.6} & \textbf{65.8} & 51.5 \\
					
			\hline
		\end{tabular}
	}
	\begin{minipage}{0.9\textwidth}\footnotesize
		\vspace{5pt}
		$\dagger$ The used checkpoint has already been trained with various data augmentations, such as instance-level rotation and scaling and therefore performs better than noted in the original paper. See \url{https://github.com/xinge008/Cylinder3D/issues/88}
	\end{minipage}
	\label{tab:results}
\end{table*}

\textsc{Table} \ref{tab:results} shows the direct comparison of the provided checkpoints with the same models but retrained with our augmentation pipeline. As KPConv \cite{thomas2019kpconv} did not provide a pretrained checkpoint we trained both the baseline and our version from scratch for a fair comparison.\\ 

In all three network categories our augmentation pipeline improves the performance without any changes to the network, hyper-parameters or training time. The class-wise comparison shows a consistent improvement for the underrepresented classes as we are forcing the networks to see these classes more often due to our injection method. We also see an improvement for classes which are not explicitly injected but mixed out of their previous context by our fusion module. Our first principle approach, to focus on the underlying lidar sensor capabilities enables us to augment the lidar data while keeping it structurally intact. Thus our approach can be applied to all kinds of semantic segmentation model for lidar data, as the augmented point clouds are structurally the same as the original raw data. 

\subsection{Benchmark Results}
\begin{table*}[!htbp]
	\caption{\textbf{Results of Cylinder3D with and without our Augmentation Pipeline on the hidden SemanticKITTI Test Set}\\}
	\resizebox{\textwidth}{!}{
		\begin{tabular}{c|c|c|c|c|c|c|c|c|c|c|c|c|c|c|c|c|c|c|c|c}
			\hline
			\textbf{Methods} & \textbf{mIoU} & \rotatebox[origin=c]{90}{car} & \rotatebox[origin=c]{90}{bicycle} & \rotatebox[origin=c]{90}{motorcycle}& \rotatebox[origin=c]{90}{truck}& \rotatebox[origin=c]{90}{other-vehicle} & \rotatebox[origin=c]{90}{person} & \rotatebox[origin=c]{90}{bicyclist} & \rotatebox[origin=c]{90}{motorcyclist} & \rotatebox[origin=c]{90}{road} & \rotatebox[origin=c]{90}{parking} & \rotatebox[origin=c]{90}{sidewalk} & \rotatebox[origin=c]{90}{other-ground} & \rotatebox[origin=c]{90}{building}& \rotatebox[origin=c]{90}{fence}& \rotatebox[origin=c]{90}{vegetation}& \rotatebox[origin=c]{90}{trunk}& \rotatebox[origin=c]{90}{terrain}& \rotatebox[origin=c]{90}{pole}& \rotatebox[origin=c]{90}{traffic-sign}\\
			
			\hline
			Cylinder3D \cite{zhou2020cylinder3d}          & 63.9          & 96.7 & 60.3          & 57.4          & 43.2          & 49.6          & 70.0          & 65.1          & \textbf{12.0}          & \textbf{91.6} & 64.6          & \textbf{76.0} & \textbf{24.3} & 90.0          & 63.4 &  \textbf{84.8}          & 70.7          & 67.5          & 62.1          & 64.0          \\
			Cylinder3D \cite{zhou2020cylinder3d} + Ours & \textbf{65.4} & \textbf{96.8} & \textbf{61.4} & \textbf{63.1} & \textbf{54.0} & \textbf{58.7} & \textbf{71.3} & \textbf{67.3} & 11.7 & 90.8 & \textbf{66.0} & 74.3 & 12.0 & \textbf{90.6} & \textbf{64.7} & \textbf{84.8} &  \textbf{73.0} & \textbf{68.4} & \textbf{64.5} & \textbf{69.1}\\
			\hline
		\end{tabular}
	}
	\label{tab:results_test}
\end{table*}
To evaluate our method further, we inferred our retrained Cylinder3D on the hidden test set of SemanticKITTI \cite{behley2019semantickitti} and compared our results to the provided checkpoint of the official repository. Please note, that we did not apply any test time augmentations or hyper-parameter tuning for both results. It is also worth mentioning, that the open sourced Cylinder3D repository is not the complete network described in \cite{zhou2020cylinder3d}, but a reduced version\footnote{https://github.com/xinge008/Cylinder3D/issues/23}. As we want to demonstrate our augmentation methods and not the capabilities of the used network, we did not re-implement the missing parts but took the code as given by the authors. \textsc{Table} \ref{tab:results_test} shows, that with our augmentation pipeline the retrained model outperforms the original checkpoint in 14 out of 19 classes, marked with bold text in the table, while they are on par for one more class.

\subsection{Ablation Study}
We compiled an ablation study of the individual augmentation components listed in \textsc{Section} \ref{section:method}. We want to show the influence of each part of our augmentation pipeline.
\begin{table}[!htbp]
	\caption{\textbf{Ablation Study using the Cylinder3D Network on the SemanticKITTI validation set}}
	\begin{tabular}{cccc|c}
		\hline
		Baseline & Global Augs. & Inject & Fusion & mIoU \\
		\hline
		\checkmark &            &            &            & 65.18 \\
		\checkmark & \checkmark &            &            & 65.73 \\
		\checkmark & \checkmark & \checkmark &            & 66.97 \\
		\checkmark & \checkmark &            & \checkmark & 67.29 \\
		\checkmark & \checkmark & \checkmark & \checkmark & 67.92 \\
		\hline
	\end{tabular}
	\centering
	\label{tab:ablation}
\end{table}
The baseline in \textsc{Table} \ref{tab:ablation} is a Cylinder3D network trained without any augmentations. Global augmentations improve the final mean Intersection over Union (mIoU) score only slightly. We see a much higher jump in the final segmentation quality when we additionally use either our fusion module or the injection module. The highest performance gain with a total of $+7.44$ mIoU points can be reached by using all augmentation modules together.

\subsection{Data Sparsity}
\begin{table}[!htbp]
	\caption{\textbf{Performance on reduced data.} Networks trained with artificially reduced subsets of the training data.}
		
	\begin{tabular}{ccccc}
		\hline
		Data  & Baseline & + Ours & $\Delta abs.$ & $\Delta rel.$ \\
		\hline
		
		100\% & 60.36 & 67.16 & +6.80 & +11.27\% \\
		
		50\% & 57.55 & 64.94 & +7.39 & +12.84\% \\
		
		10\% & 53.90 & 63.98 & +10.08 & +18.70\% \\
		
		1\% &  37.32 & 50.86 & +13.54 & +36.28\% \\
		\hline
	\end{tabular}
	\label{tab:dataset-reduction}
	\centering
\end{table}

We investigated, if our augmented data is on par with real raw lidar data. 
As our pipeline creates new scenes by recombining multiple parts of the present data we wanted to compare remixed augmented data to additional new data. In \textsc{Table} \ref{tab:dataset-reduction} we artificially decrease the dataset by keeping every second, every tenth and every one hundredths data sample in the scene database as well as our injection database, without any data mining for diverse scenes. We train Cylinder3D \cite{zhou2020cylinder3d} networks from scratch on these subsets and evaluate them on the full validation set. We observe that our augmentation methods improve the baseline on all data subsets. Particularly noteworthy is the fact, that we noticeably outperform the baseline trained on the full dataset with only 10\% of the original data by 4.2 mIoU by applying our augmentation pipeline. This indicates that the generated scenes we create are more valuable than simply more raw data, as even a ten fold increase of training data causes less improvement than our pipeline on the reduced small dataset.

\section{Conclusion and Future Work}
In this work, we presented methods to artificially compensate for imbalanced datasets. We use a lidar-centric approach throughout, which enables us to augment data in a structure very similar to the lidar sensors' working principle. This conserved structure enables the use of these methods with all kinds of neural networks, including those relying on the lidar structure.
In our experiments, we have shown that the augmentation methods outlined in this paper improve the performance of multiple lidar semantic segmentation models.
In addition, we have shown that our method can be used to successfully generalize networks with small datasets. This is particularly interesting because annotating data is time consuming and costly.
In the future we intend to investigate the performance of our augmentation pipeline on more dataset and on other tasks using lidar scanners, such as instance and panoptic segmentation as well as object detection.




{\small
	\bibliographystyle{ieee_fullname}
	\bibliography{Bibliography}
}

\end{document}